\documentclass[letterpaper, 10 pt, conference]{ieeeconf}  

\IEEEoverridecommandlockouts                              

\usepackage{hyperref}
\usepackage{graphicx}
\usepackage{amsmath}
\usepackage{cite}

\title{\LARGE \bf
One ACT Play: Single Demonstration Behavior Cloning with Action Chunking Transformers}

\author{Abraham George$^{1}$ and Amir Barati Farimani$^{1}$
\thanks{$^{1}$With the Department of Mechanical Engineering,
        Carnegie Mellon University 
        {\tt\small \{aigeorge, afariman\} @andrew.cmu.edu}}%
}

\begin{document}

\maketitle
\thispagestyle{empty}
\pagestyle{empty}

\begin{abstract}
Learning from human demonstrations (behavior cloning) is a cornerstone of robot learning. However, most behavior cloning algorithms require a large number of demonstrations to learn a task, especially for general tasks that have a large variety of initial conditions. Humans, however, can learn to complete tasks, even complex ones, after only seeing one or two demonstrations. Our work seeks to emulate this ability, using behavior cloning to learn a task given only a single human demonstration. We achieve this goal by using linear transforms to augment the single demonstration, generating a set of trajectories for a wide range of initial conditions. With these demonstrations, we are able to train a behavior cloning agent to successfully complete three block manipulation tasks. Additionally, we developed a novel addition to the temporal ensembling method used by action chunking agents during inference. By incorporating the standard deviation of the action predictions into the ensembling method, our approach is more robust to unforeseen changes in the environment, resulting in significant performance improvements.

\end{abstract}


\section{INTRODUCTION}
\begin{figure}[thpb]
      \centering
      \includegraphics[width=\linewidth]{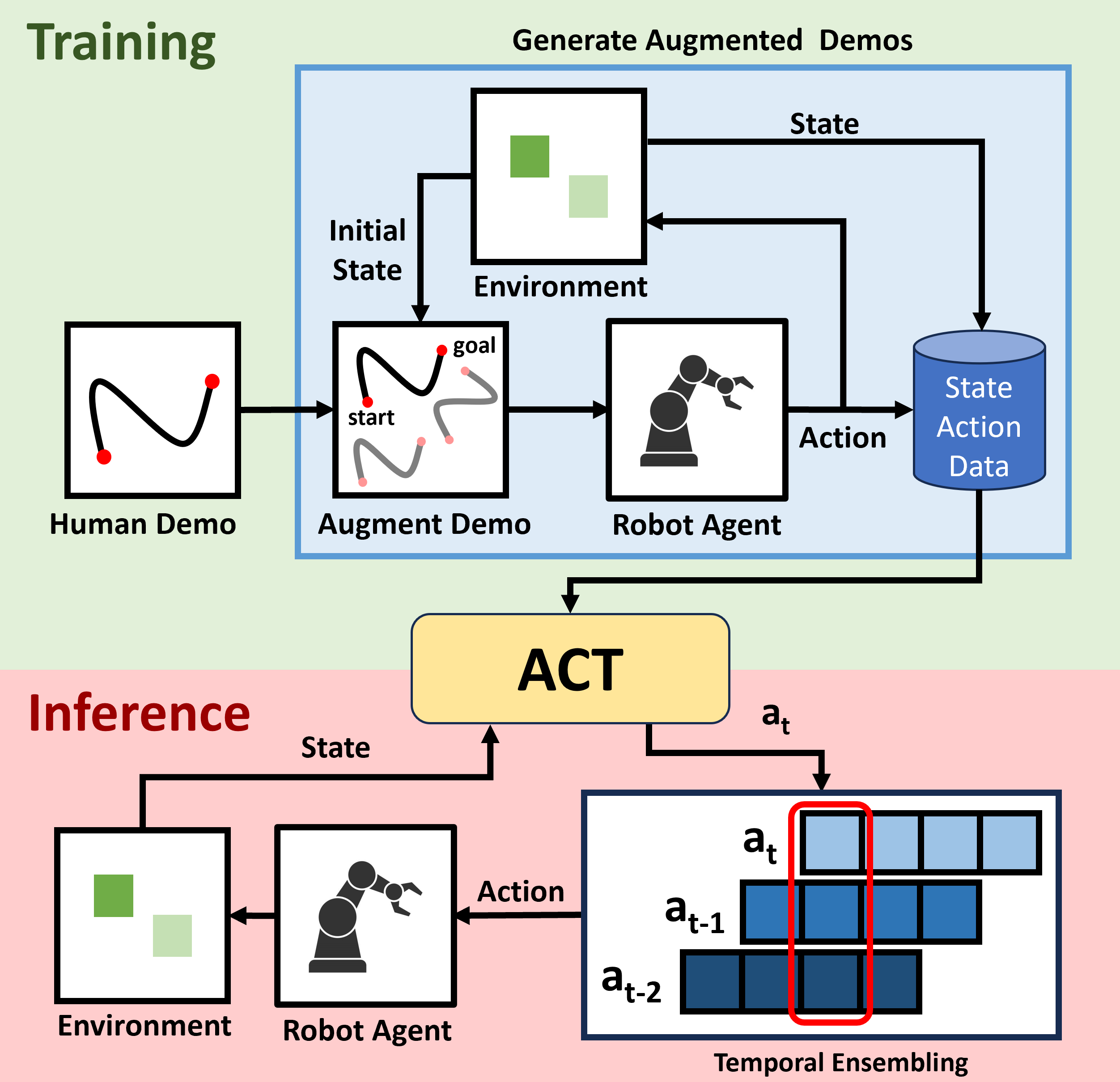}
      \caption{\label{fig:overview} Diagram outlining our method. A single demonstration trajectory is augmented then replayed, and the resulting state-action observations are used to train our behavior cloning agent (ACT). At inference time, the environment state is passed to the agent, which returns an action chunk, $a_t$. The action chunk is combined with prior action chunks via temporal ensembling, and the resulting action is executed.}
      \label{figurelabel}
  \end{figure}

Behavior cloning, or the process of teaching an agent to mimic the actions of a human in order to complete a task, is a key aspect of robotic learning. It has been used to teach agents to complete tasks ranging from driving cars \cite{Pomerleau-1989-15721, bojarski2016end, 8855753} to playing video games \cite{vinyals2019grandmaster}, to robotic locomotion \cite{torabi2018behavioral} and
complex manipulation \cite{Mandlekar2021WhatMI, shafiullah2022behavior, 9289148}. However, behavior cloning has many challenges, including compounding errors that lead to unpredictable out-of-distribution performance \cite{ross2011, Tu2021OnTS}, and sample inefficiency \cite{10.1145/3054912}. Although much progress has been made recently in addressing these issues, mitigating the problems of compounding errors and unpredictable performance using strategies such as action chunking \cite{lai2022action} and increasing sample efficiency through both data set augmentation and improved network architecture \cite{8971711, wang2023diffusion}, the limitations, especially for sample efficiency, persist. In particular, the issue of poor sample efficiency means that behavior cloning agents require many demonstrations, often in the hundreds, to learn tasks that a human could master with only a single demonstration \cite{8461249}. Recent work in the related field of reinforcement learning (RL) with human demonstrations has addressed the issue of sample efficiency by augmenting a single demonstration using simple linear transforms, then autonomously replaying the augmented demonstrations and observing the resulting states \cite{10161119}. In this work, we explore applying a similar augmentation method in a behavior cloning setting to develop a method to learn a task given only a single human demonstration. An outline of our method can be found in Figure \ref{fig:overview}. 

Because the agent's training data originates from a single demonstration, the agent is only able to learn on a small portion of the task's state space. As such, the ability to generalize to unseen states and recover from poor predictions is vital for our behavior cloning algorithm. Therefore, we chose to base our method on Action Chunking with Transformers (ACT), whose use of a convolutional autoencoder (CVAE) increases generalizability, and whose action chunking and ensembling methods make the agent resistant to occasional poor actions \cite{zhao2023learning}. However, we found that the original action ensembling method (a weighted average of actions predicted for the current time step, each predicted at a different prior time step) was not suited for the block manipulation tasks we evaluated our method on. If the task does not go as the agent expected, some of the previous action predictions may become erroneous, corrupting the cumulative action. To address this issue, we introduce a heuristic, based on the standard deviation of the predicted actions, to estimate if the action predictions are in agreement. If they are not, we alter the ensembling method to ignore older action predictions, and if the disagreement is very large, we temporarily suspend action ensembling, instead replaying a single action chunk. Finally, we evaluated both our single demonstration behavior cloning method and novel ensembling method on three block manipulation tasks: moving a block across a table to a goal location (push), moving a block to a goal location above the table (pick-and-place), and stacking two blocks, in the correct order, at a specified location on a table (stack).

\section{Related Works}

\subsection{Behavior Cloning}
Behavior cloning uses demonstrations to determine an agent's actions by having the agent replicate expert examples \cite{Pomerleau-1989-15721}. This task can be accomplished through machine learning using a supervised learning approach such as classification \cite{ratliff2007}. These forms of behavior cloning have proven effective on complex tasks such as biped locomotion \cite{nakanishi2004}, but they require large data sets and do not function well in environments outside of the range of examples they trained on \cite{reichlin2022back}. Data set Aggregation (DAgger) addresses some of theses issues by augmenting the learned policy with expert demonstrations collected throughout the training process, interconnecting the learned and expert policies \cite{ross2011}. However, DAgger can be difficult to implement because it requires the use of expert examples collected throughout the duration of training. 

\subsection{Single Demonstration Reinforcement Learning}
Multiple methods have been developed for learning from a single demonstration in the field of reinforcement learning, where the addition of a single demonstration is primarily used to overcome the exploration problem \cite{pmlr-v80-kang18a, subramanian2016exploration, brys2015reinforcement}. One strategy is to use a single demonstration through curriculum learning, which trains the agent on progressively larger sub-sections of the desired task \cite{10.1145/1553374.1553380}. \cite{salimans2018learning} used this method to beat Montezuma's Revenge, an Atari game with a long-horizon for reward, with a single demonstration. By resetting the game to states along the demonstration path, starting near the end of the demonstration and progressively moving further and further back, a curriculum of steadily increasing difficulty was created that enabled PPO \cite{schulman2017proximal} to achieve a SOTA score on the game. Similar curriculum methods have been used by \cite{florensa2018reverse} to help train robotic tasks. Another approach to training reinforcement learning agents using demonstrations is to directly add the demonstrations to the replay buffer \cite{lipton2016, nair2018}. This approach was used by \cite{10161119}, in combination with curriculum learning, to train an RL agent. This work only used a single demonstration, augmented to form 5000 'human-like' demonstrations by linearly scaling the demonstration trajectory, showing that a simple augmentation method could result in significant improvements in performance. Although learning from a human demonstration, the agent often learned a policy significantly different from that shown, illustrating the creativity of reinforcement learning \cite{vargas2019creativity}.

\subsection{Action Chunking with Transformers}
Action Chunking with Transformers (ACT) is a behavior cloning algorithm that uses a Conditional Variational Autoencoder (CVAE) to model diverse scenes, combined with a transformer to be able to predict action sequences (chunks) given a multimodal input \cite{zhao2023learning}. By casting these actions as goal states for the manipulator to reach, a temporal aggregation method can be used to combine predicted actions from multiple previous time steps through a weighted average. By combining multiple predictions, this approach mitigates the problems of compounding errors and unpredictable responses to out-of-distribution states. Although erroneous actions may still be chosen by the model, correct actions, predicted during previous time steps, will provide a moderating influence on the final action. Additionally, the transformer network structure \cite{46201} allows for a wide range of multi-model inputs, such as text prompts, further improving the robustness of the ACT framework \cite{bharadhwaj2023roboagent}.

\section{METHODS}

\subsection{Human Demonstration Collection}
The single human demonstration used by our behavior cloning method was collected using an Oculus Quest 2 virtual reality headset, running the unity-based teleoperation environment developed by \cite{george2023openvr}. This is the same approach as was taken by \cite{10161119}. The VR environment shows the user a first-person view of the task they need to complete along with a virtual Franka-Emika Panda end-effector, which the user controls using the Oculus's hand-held controller. For example, to demonstrate a pick-and-place task, the user is shown the block to move, along with a transparent goal block, highlighting the region where the block is to be placed. The demonstration is done entirely in simulation, using Unity's physics engine, which has slightly different dynamics than the pybullet simulator \cite{coumans2019} used for testing, or the hardware system used for validation. The user's actions are recorded while completing the desired task, creating a trajectory file that will be used by the augmentation method to generate additional demonstrations. 

\subsection{Demonstration Augmentation}
Behavior cloning algorithms treat the control problem as a supervised learning task, imitating a human's action (or series of actions) for a given state. Therefore, to learn a task the agent must be trained on a wide range of demonstrations from across the task's state space so that the agent can extrapolate the correct action for any given state. However, our approach only involves a single demonstration. As such, we require an augmentation method to turn that single demonstration into a collection of demonstrations that covers a sizeable portion of the task's state space. To accomplish this, we turn to the linear scaling method developed by \cite{10161119}. However, since our task is behavior cloning, not reinforcement learning, we have to take more care about the quality of the demonstration we collect. As such, rather than scaling and shifting each axis, we apply a linear transform to the trajectory consisting of rotation, translation, and uniform scaling, which results in less distortion in the generated trajectory. 

To generate a new trajectory using our recorded demonstration trajectory, we first generate a random start and goal location. A rotation matrix is then calculated such that the vector from the recorded start location to the recorded goal location, once rotated, will align with the vector from the generated start to the generated goal locations. This constraint leaves a degree of freedom, as the transformed vector is free to rotate about its axis. We use this degree of freedom to stipulate that the z-axis of the transformed vector should be aligned with the world frame's z-axis. This constraint ensures that "up" is preserved in the augmented demos, which is important due to gravity. The two constraints for the rotation matrix are shown below:

\begin{equation}
    r_\Delta = r_g - r_s, \quad g_\Delta = g_g - g_s 
\end{equation}
\begin{equation}
    \frac{R\,r_\Delta}{\|R\,r_\Delta\|} = \frac{g_\Delta}{\|g_\Delta\|}, \quad
    \frac{\text{Proj}_{(R\,\hat{z})}(P)}{\|\text{Proj}_{(R\,\hat{z})}(P)\|} = \frac{\text{Proj}_{(\hat{z})}(P)}{\|\text{Proj}_{(\hat{z})}(P)\|}
\end{equation}

Where $r_g$ is the recorded goal, $r_s$ is the recorded start, $g_g$ is the generated goal, $g_s$ is the generated start, $R$ is the rotation matrix, $P$ is a plane with normal vector $g_\Delta$, $\hat{z}$ is a vertical unit vector, and $\text{Proj}_{(a)}(B)$ means the projection of vector $a$ onto plane $B$.

Next, a scaling factor is calculated so that the distance between the recorded start and recorded end location matches the distance between the generated start and generated end locations. Finally, a translation matrix is calculated to make the rotated, scaled, and translated recorded start location match the generated start location. These transforms, along with the rotation matrix from above, are combined to give a final transform, as shown below:

\begin{equation}
    s = \frac{||g_\Delta||}{||r_\Delta||},\quad t = (g_s - R\,r_s)
\end{equation}
\begin{equation}
    T = 
    \begin{bmatrix}
    s & 0 & 0 & t_x \\
    0 & s & 0 & t_y \\
    0 & 0 & s & t_z \\
    0 & 0 & 0 & 1
    \end{bmatrix} 
    \begin{bmatrix}
    R & 0 \\
    0 & 1 
    \end{bmatrix} 
\end{equation}

Where $T$ is the final linear transform. 

Once the linear transform for a given environment is calculated, the recorded points from the single demo are transformed, and the resulting trajectory is replayed using a proportional controller. The augmentation method assumes the task can be linearly transformed given a start and goal location. If the task is more complex, it can instead be decomposed into multiple sub-tasks, each of which can be linearly transformed. For example, for block stacking, the trajectory can be split into multiple sub-trajectories (move block one, then move block two, etc.), and each sub-trajectory can then be warped independently. 

When the agent replays the generated trajectory, state-action information is collected as if the agent were human-controlled. If the augmented demonstration is unsuccessful at completing the task, it is discarded. Because this method disposes of bad demonstrations, the requirement for effectiveness of the augmented trajectories is mainly dependent on the expense of playing back trajectories; if replaying trajectories is relatively inexpensive (such as in simulation) then low accuracy can be mitigated by increasing the volume of simulations.
 
\subsection{Learning Architecture}
\begin{figure}[thpb]
      \centering
      \includegraphics[width=\linewidth]{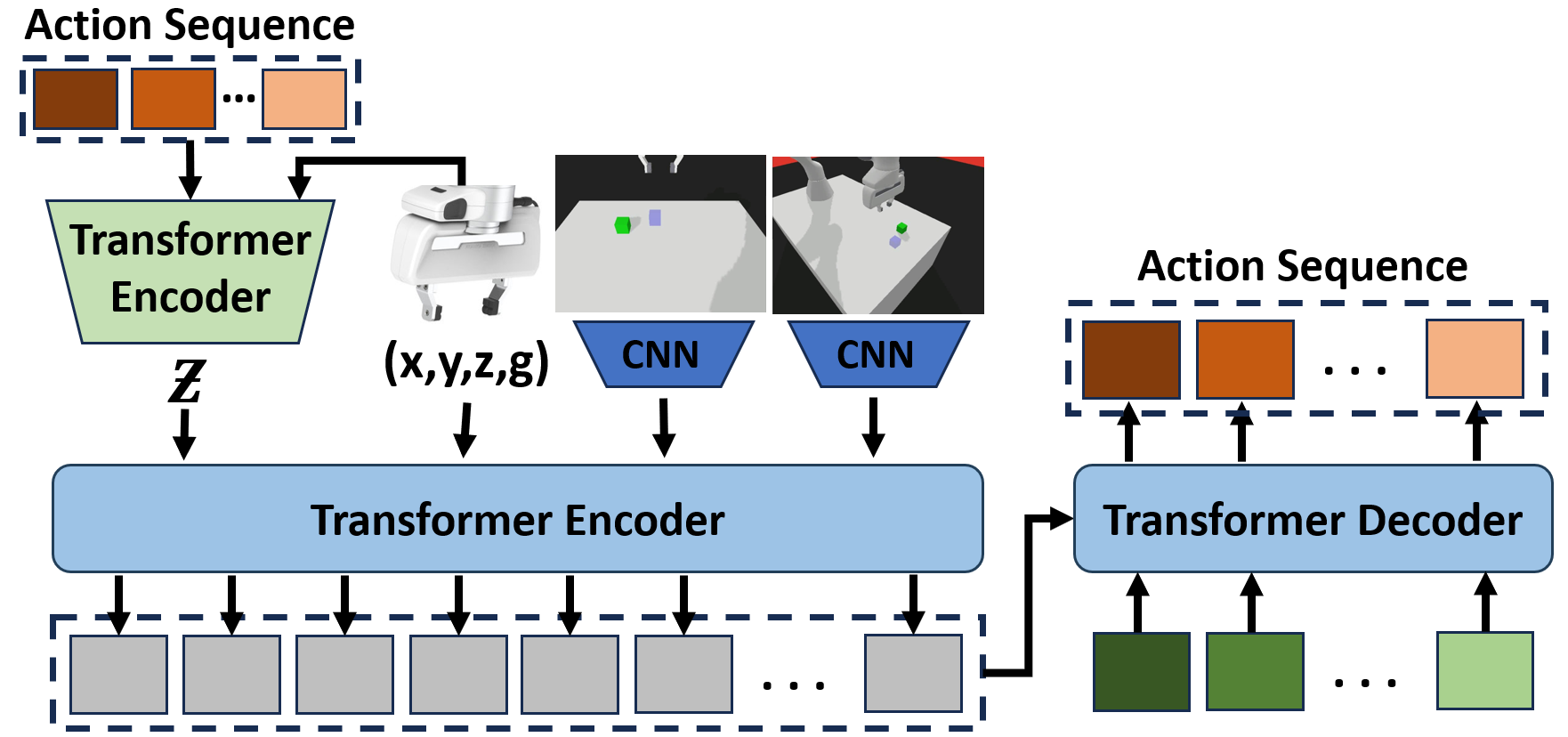}
      \caption{\label{fig:network} Diagram of the network structure used for the behavior cloning model, based on the ACT framework \cite{zhao2023learning}. The network is trained as a conditional variational auto-encoder (CVAE) on the action sequence. During training, a style variable $Z$ is calculated by a transformer encoder from the action sequence and the agent's position. During inference, this encoder is removed and $Z$ is set to $0$.}
      \label{figurelabel}
  \end{figure}

To develop efficient policy architectures for robot manipulation, especially with the limited variety of data produced through single demo augmentation, we must address several key objectives. First, we need a policy that can generalize well to unseen states outside of the training distribution. Additionally, the policy should be robust to a small number of poor predictions, so that a few incorrect choices (potentially caused by out-of-distribution states) do not derail the agent's overall policy. For these reasons, we based our method on Action Chunking with Transformers \cite{zhao2023learning}. Our network structure is very similar to the original ACT network, except for minor alterations due to differences in state representations. We chose to control the location and gripper width of a single arm with a parallel plate gripper, whereas the original work controlled joint angles of a 2 arm setup. Similarly to ACT, we chose to use a pixel-level observation space, removing the need for an additional vision system to extract permanent state information \cite{dikshit2023robochop}. A diagram of our network structure can be seen in Figure \ref{fig:network}.

Additionally, we have altered the temporal ensembling method to better account for dynamic and multimodal environmental states. The original ACT algorithm determined its action at inference time by calculating a weighted average of actions (goal end-effector positions) it previously predicted for that time step, with weights of the form $e^{-kt}$, where $t$ is the time since the prediction was made. Although this strategy works well, decreasing action noise and limiting the effect of erroneous actions due to out-of-distribution states, the use of a weighted average makes the implicit assumption that the predicted actions consist of a correct action, with symmetric noise added. Under this assumption, using an average action will lead the model to choose the underlying correct action. However, if there are multiple possible approaches, the predicted actions can form a multi-modal distribution, clustered around different "correct" actions. In this situation, a weighted average will lead to an in-between action, part of neither approach, and likely a poor option. This issue is exacerbated in non-stationary environments and environments with distinct state changes, where earlier predictions can be quite bad. For example, in block manipulation tasks, the choice of action at a future time step is heavily dependent on whether the agent thinks it will have successfully grasped the block at that time. If this prediction changes during execution (such as if the gripper motion is slower than expected) then the predicted action distribution will be bi-modal (some actions assuming the block is gripped, some actions assuming it is not), causing the average action to be a poor choice for either situation. 

Our implementation of temporal aggregation addresses the issue of multi-modal action distributions by dynamically adjusting the temperature value, $k$, for the exponentially decaying weights used in the weighted average. If the action distribution is clustered around a value, then we assume that the predicted state has been consistent, and therefore using the average of all of the predicted actions would be effective. However, if the distribution of actions is highly variant, then we assume that the predicted state has not been static and that earlier action predictions may be erroneous and should be ignored. Given this relationship, we choose $k$ to be proportional to the standard deviation of the distribution. If the actions are widely distributed, then $k$ will be large, causing the agent to ignore previous action predictions, and if the predicted actions are tightly clustered, then $k$ will be small, causing the agent to incorporate past actions more equally. Because our action is comprised of two modalities, end-effector position and gripper width, two $k$ values are calculated. Because the position is a vector, we use the $L_\infty$ norm of its standard deviation for the positional $k$.

\begin{equation}
    k_{g} = \beta\sigma(a_{g}), \quad k_{p} = \beta||\sigma(a_p)||_{\infty}
\end{equation}

Where $k_g$, $a_g$ and $k_p$, $a_p$ are the temperature constants and predicted actions for the gripper width and end-effector positions, respectively, and $\beta$ is a proportionality constant.

This approach addresses drift in the action space, where the difference between the expected and resulting state leads to differing action predictions. However, if this drift is severe, or if the action predictions are oscillating between multiple different action policies, $k$ becomes very large, rendering action chunking moot and reinstating the issues of compounding errors and out-of-distribution states that action chunking was designed to address. In this case, we prefer to keep action chunking and instead eliminate temporal ensembling. If $k$ goes above a specified cutoff, we directly sample the next $n$ time steps directly from the current action chunk prediction, where $n$ is a hyperparameter (we chose $n$ to be one-half of the action chunk length). This allows the agent to execute a coherent series of actions, which ideally removes the agent from the state that was causing temporal ensembling trouble. Because our method is based on standard deviation, we use the original ensembling method until the fifth time step of a run to ensure a sufficient sample size of action predictions.

\section{EXPERIMENTAL EVALUATION}

\subsection{Simulation Environment}

\begin{figure}[thpb]
      \centering
      \includegraphics[width=\linewidth]{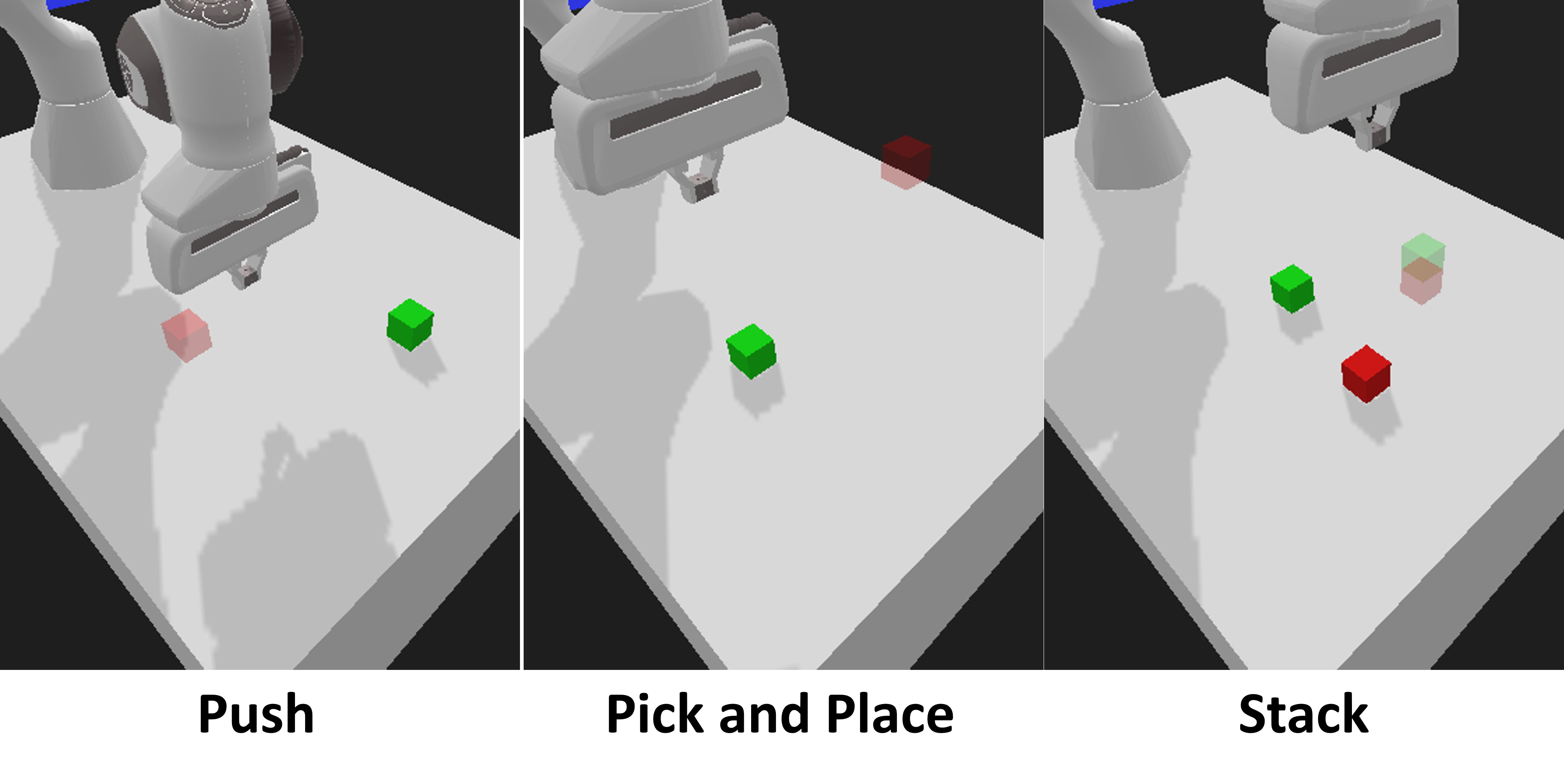}
      \caption{\label{fig:gym_env}Panda-Gym environments used for evaluation. The block(s) to be stacked are shown as opaque cubes, while the goal locations are shown as transparent cubes.}
      \label{figurelabel}
  \end{figure}

 We assessed the efficacy of our singular demonstration behavior cloning approach on three Panda-Gym tasks developed by Gallouedec et al.: block pushing, block pick-and-place, and block stacking, as illustrated in Figure \ref{fig:gym_env} \cite{gallouedec2021}. These Panda-Gym tasks use a seven-degree-of-freedom Franka Emika Panda robot arm to manipulate one or more 4 cm blocks in a PyBullet simulation. Throughout the experiments, the gripper's orientation remains fixed, rendering the end effector a four-degree-of-freedom system, consisting of position and gripper width. Task success is based on the distance between the current cube position (depicted as the opaque cube in Figure \ref{fig:gym_env}) and the desired cube location (represented by the transparent cube in Figure \ref{fig:gym_env}), with a success cutoff of 5 cm for the push and pick-and-place tasks, and 4 cm for both cubes in the stacking task. To make the push task more intuitive for human teleoperation, we modify the existing implementation to permit fingertip mobility, allowing the user to use the gripper to complete the task. This change effectively transforms the push task into a two-dimensional pick-and-place task. 

\begin{figure}[thpb]
      \centering
      \includegraphics[width=\linewidth]{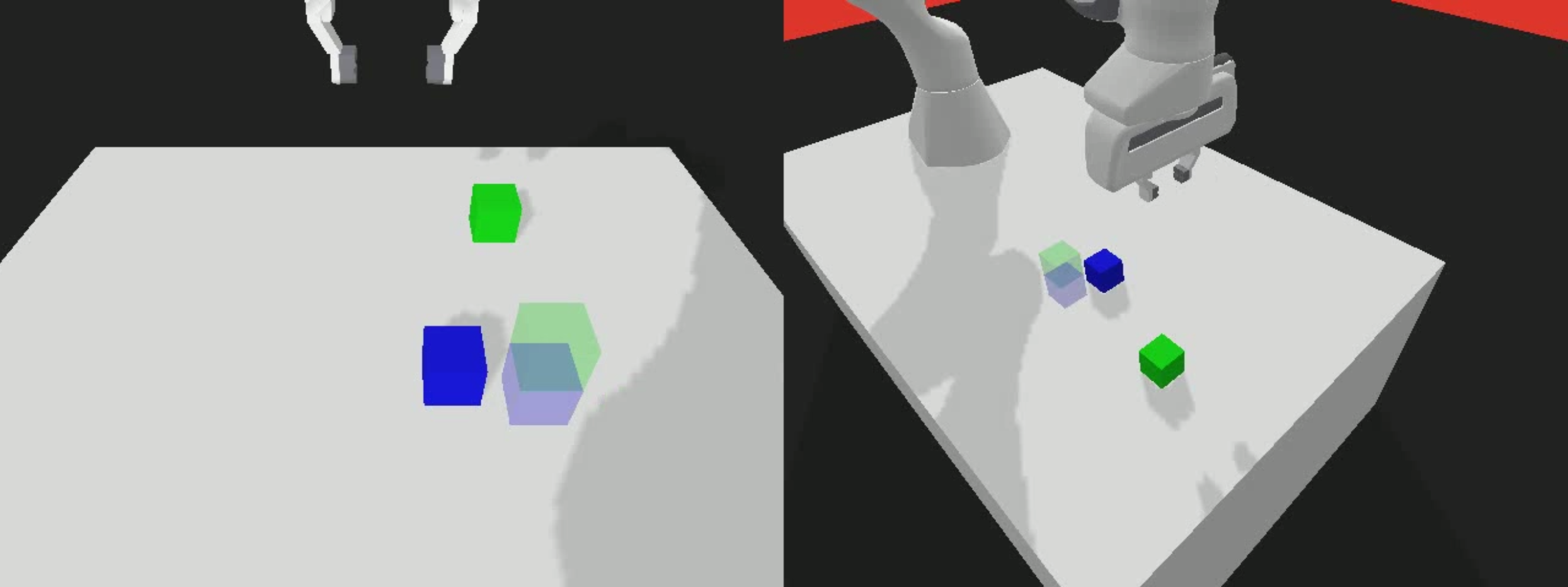}
      \caption{\label{fig:sim_view}Example views from the two cameras used in simulation experiments. On the left is a view from the end-effector camera, and on the right a view from the isometric camera.}
  \end{figure}

The behavior cloning agent observes the scene using two virtual cameras, one mounted on the end-effector and another located in the top corner of the environment, providing an isometric view of the scene (see Figure \ref{fig:sim_view}). In addition to the two cameras, the agent observes the state of the robot's end-effector, consisting of the x, y, and z position of the center of the gripper, and the gripper's width (x,y,z,g).

\subsection{Results}
\subsubsection{Single Demonstration Behavior Cloning}
\begin{figure}[thpb]
      \centering
      \includegraphics[width=\linewidth]{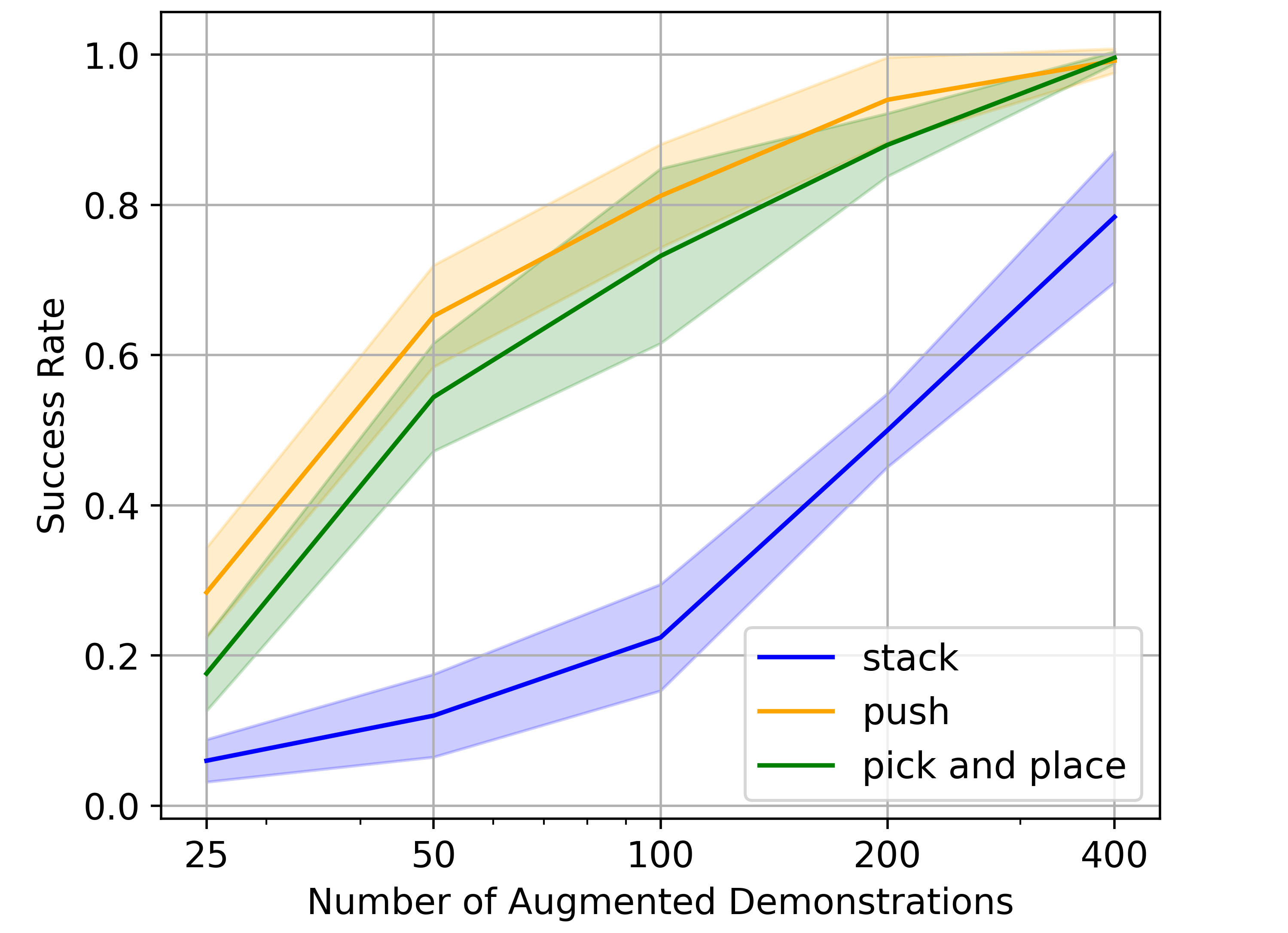}
      \caption{\label{fig:eval} Success rates of our behavior cloning method for different numbers of augmented demonstrations used in training. The method was evaluated on the push, pick and place, and stack tasks; the experiments were run in simulation. The shaded region on the graph shows one standard deviation from the mean.}
      \label{figurelabel}
  \end{figure}

To examine the effectiveness of learning a policy from a single demonstration using our augmentation method and our variation of Action Chunking with Transforms, we trained our behavior cloning agent on the three evaluation tasks (push, pick and place, and stack) with a varying number of augmented demonstrations (25, 50, 100, 200, and 400). The agent's success rates, after training, are shown in Figure \ref{fig:eval}. Our results show that using a single demonstration, augmented with linear transforms, a behavior cloning agent is able to learn all three block manipulation tasks, with a nearly perfect success rate for push and pick and place, and an impressive 78.4\% success rate for the more complicated stack task. Additionally, these results show a significant increase in performance as the number of augmented demonstrations increases. This relationship was expected since more augmentations mean more demonstrations, increasing the variety of experience the BC agent is exposed to, leading to a more complete coverage of the state space, and an according decrease in out-of-distribution states. Additionally, our results show the number of augmented demonstrations needed to learn a task is proportional to the complexity of the task, which is in line with observations made for similar BC tasks that directly used demonstrations \cite{Mandlekar2021WhatMI}. 

\subsubsection{Temporal Ensembling} 
To examine the effectiveness of our temporal ensembling method, we re-ran our behavior cloning experiments with the original temporal ensembling method proposed by \cite{zhao2023learning}, using an exponentially decaying weighted average with a constant $k$. The results from this experiment can be seen in Table 1. Compared with our ensembling method, we observe that the baseline performance is slightly worse for the push and pick and place tasks, and significantly worse for the stack task. Because the stack task is the most complex task with the longest action horizon, it is more likely to suffer from drift or multi-modality in its action predictions, making our changes to address these issues more relevant. 

Our ensembling method has two main components: a dynamic heat constant, $k$, based on standard deviation and the temporary suspension of temporal ensembling, instead using pure action chunking, when the ensembled actions become too varied. To examine the impact of each of these aspects individually, we re-ran our experiments with only the dynamic heat constant and only the suspension of temporal ensembling (fixed heat constant). Because the effect of the dynamic heat constant is largely dependent on the proportionality constant, beta, used to calculate $k$ from the standard deviation, we ran experiments with beta equal to 1, 0.5, and 0.25 for both only a dynamic $k$ and the combined method. The results can be found in Table 1.

\begin{table}[]
\setlength{\tabcolsep}{4.6pt}
\label{table}
\caption{Temporal Ensembling Method Ablation Study}
\begin{tabular}{rccccccccc}
\hline
\multicolumn{10}{|c|}{Baseline Success Rate (\%)}                                                                                                                                                                                                              \\ \hline
\multicolumn{1}{|l|}{\#Demos} & \multicolumn{3}{c|}{Push}                                            & \multicolumn{3}{c|}{Pick and Place}                                  & \multicolumn{3}{c|}{Stack}                                           \\ \hline
\multicolumn{1}{|r|}{25}       & \multicolumn{3}{c|}{29.2}                                               & \multicolumn{3}{c|}{22.4}                                               & \multicolumn{3}{c|}{5.6}                                               \\
\multicolumn{1}{|r|}{50}       & \multicolumn{3}{c|}{60.8}                                               & \multicolumn{3}{c|}{42.8}                                               & \multicolumn{3}{c|}{7.6}                                               \\
\multicolumn{1}{|r|}{100}      & \multicolumn{3}{c|}{78.0}                                               & \multicolumn{3}{c|}{70.0}                                               & \multicolumn{3}{c|}{22.8}                                               \\
\multicolumn{1}{|r|}{200}      & \multicolumn{3}{c|}{92.0}                                               & \multicolumn{3}{c|}{84.0}                                               & \multicolumn{3}{c|}{44.4}                                               \\
\multicolumn{1}{|r|}{400}      & \multicolumn{3}{c|}{98.8}                                               & \multicolumn{3}{c|}{98.8}                                               & \multicolumn{3}{c|}{60.8}                                               \\ \hline
\multicolumn{1}{l}{}           & \multicolumn{1}{l}{} & \multicolumn{1}{l}{} & \multicolumn{1}{l}{}   & \multicolumn{1}{l}{} & \multicolumn{1}{l}{} & \multicolumn{1}{l}{}   & \multicolumn{1}{l}{} & \multicolumn{1}{l}{} & \multicolumn{1}{l}{}   \\ \hline

\multicolumn{10}{|c|}{Only Resetting Success Rate (\%)}                                                                                                                                                                                                              \\ \hline
\multicolumn{1}{|l|}{\#Demos} & \multicolumn{3}{c|}{Push}                                            & \multicolumn{3}{c|}{Pick and Place}                                  & \multicolumn{3}{c|}{Stack}                                           \\ \hline
\multicolumn{1}{|r|}{25}       & \multicolumn{3}{c|}{34.0}                                               & \multicolumn{3}{c|}{12.8}                                               & \multicolumn{3}{c|}{6.4}                                               \\
\multicolumn{1}{|r|}{50}       & \multicolumn{3}{c|}{52.4}                                               & \multicolumn{3}{c|}{44.0}                                               & \multicolumn{3}{c|}{9.2}                                               \\
\multicolumn{1}{|r|}{100}      & \multicolumn{3}{c|}{81.2}                                               & \multicolumn{3}{c|}{64.0}                                               & \multicolumn{3}{c|}{18.8}                                               \\
\multicolumn{1}{|r|}{200}      & \multicolumn{3}{c|}{90.8}                                               & \multicolumn{3}{c|}{80.4}                                               & \multicolumn{3}{c|}{40.4}                                               \\
\multicolumn{1}{|r|}{400}      & \multicolumn{3}{c|}{97.2}                                               & \multicolumn{3}{c|}{89.2}                                               & \multicolumn{3}{c|}{58.4}                                               \\ \hline
\multicolumn{1}{l}{}           & \multicolumn{1}{l}{} & \multicolumn{1}{l}{} & \multicolumn{1}{l}{}   & \multicolumn{1}{l}{} & \multicolumn{1}{l}{} & \multicolumn{1}{l}{}   & \multicolumn{1}{l}{} & \multicolumn{1}{l}{} & \multicolumn{1}{l}{}   \\ \hline
\multicolumn{10}{|c|}{Only Adjusting $k$ Success Rate (\%)}                                                                                                                               
                                                                             \\ \hline

\multicolumn{1}{|l|}{\#Demos} & \multicolumn{3}{c|}{Push}                                            & \multicolumn{3}{c|}{Pick and Place}                                  & \multicolumn{3}{c|}{Stack}                                           \\ \hline
\multicolumn{1}{|r|}{$\beta$} & 1/4                  & 1/2                 & \multicolumn{1}{c|}{1} & 1/4                  & 1/2                  & \multicolumn{1}{c|}{1} & 1/4                  & 1/2                  & \multicolumn{1}{c|}{1} \\ \hline
\multicolumn{1}{|r|}{25}       & 30.8                    & 32.4                    & \multicolumn{1}{c|}{26.8} & 21.6                    & 20.4                    & \multicolumn{1}{c|}{18.0} & 5.6                    & 5.6                    & \multicolumn{1}{c|}{6.8} \\
\multicolumn{1}{|r|}{50}       & 54.0                    & 52.4                    & \multicolumn{1}{c|}{46.4} & 46.0                    & 41.2                    & \multicolumn{1}{c|}{46.0} & 14.0                    & 9.6                    & \multicolumn{1}{c|}{7.6} \\
\multicolumn{1}{|r|}{100}      & 81.2                    & 80.4                    & \multicolumn{1}{c|}{77.2} & 67.2                    & 68.0                    & \multicolumn{1}{c|}{65.6} & 17.2                    & 20.0                    & \multicolumn{1}{c|}{23.6} \\
\multicolumn{1}{|r|}{200}      & 89.2                    & 90.0                    & \multicolumn{1}{c|}{91.2} & 82.8                    & 83.6                    & \multicolumn{1}{c|}{81.2} & 42.4                    & 41.6                    & \multicolumn{1}{c|}{40.0} \\
\multicolumn{1}{|r|}{400}      & 99.2                    & 98.4                    & \multicolumn{1}{c|}{98.0} & 96.8                    & 92.8                    & \multicolumn{1}{c|}{98.8} & 67.6                    & 66.8                    & \multicolumn{1}{c|}{59.2} \\ \hline
\multicolumn{1}{l}{}           & \multicolumn{1}{l}{} & \multicolumn{1}{l}{} & \multicolumn{1}{l}{}   & \multicolumn{1}{l}{} & \multicolumn{1}{l}{} & \multicolumn{1}{l}{}   & \multicolumn{1}{l}{} & \multicolumn{1}{l}{} & \multicolumn{1}{l}{}  \\ \hline
\multicolumn{10}{|c|}{Resetting and Adjusting $k$ Success Rate (\%)}
\\ \hline

\multicolumn{1}{|l|}{\#Demos} & \multicolumn{3}{c|}{Push}                                            & \multicolumn{3}{c|}{Pick and Place}                                  & \multicolumn{3}{c|}{Stack}                                           \\ \hline
\multicolumn{1}{|r|}{$\beta$} & 1/4                  & 1/2                  & \multicolumn{1}{c|}{1} & 1/4                  & 1/2                  & \multicolumn{1}{c|}{1} & 1/4                  & 1/2                  & \multicolumn{1}{c|}{1} \\ \hline
\multicolumn{1}{|r|}{25}       & 30.4                    & 26.0                    & \multicolumn{1}{c|}{28.4} & 14.8                    & 16.0                    & \multicolumn{1}{c|}{17.6} & 6.0                    & 4.0                    & \multicolumn{1}{c|}{6.0} \\
\multicolumn{1}{|r|}{50}       & 60.4                    & 54.8                    & \multicolumn{1}{c|}{65.2} & 49.6                    & 43.2                    & \multicolumn{1}{c|}{54.4} & 10.0                    & 8.4                    & \multicolumn{1}{c|}{12.0} \\
\multicolumn{1}{|r|}{100}      & 81.6                    & 78.0                    & \multicolumn{1}{c|}{81.2} & 71.6                    & 67.6                    & \multicolumn{1}{c|}{73.2} & 21.2                    & 20.8                    & \multicolumn{1}{c|}{22.4} \\
\multicolumn{1}{|r|}{200}      & 91.2                    & 87.2                    & \multicolumn{1}{c|}{94.0} & 81.2                    & 82.8                    & \multicolumn{1}{c|}{88.0} & 41.6                    & 51.6                    & \multicolumn{1}{c|}{50.0} \\
\multicolumn{1}{|r|}{400}      & 92.8                    & 98.4                    & \multicolumn{1}{c|}{99.2} & 97.2                    & 99.2                    & \multicolumn{1}{c|}{99.6} & 68.4                    & 72.0                    & \multicolumn{1}{c|}{78.4} \\ \hline
\end{tabular}
\end{table}

Our results show that using only a dynamic $k$, with a good $\beta$ value, performs slightly better than baseline, and using only resetting performs slightly worse than baseline. However, combining the two approaches has the greatest success rate. Additionally, we found that our augmentation method is quite susceptible to the choice of $\beta$, with a $\beta$ of 0.5 performing the best when only using a dynamic $k$, and a $\beta$ of 1 performing the best when using the combined approach.

\subsection{Hardware Validation} 
\begin{figure}[thpb]
  \centering
  \includegraphics[width=\linewidth]{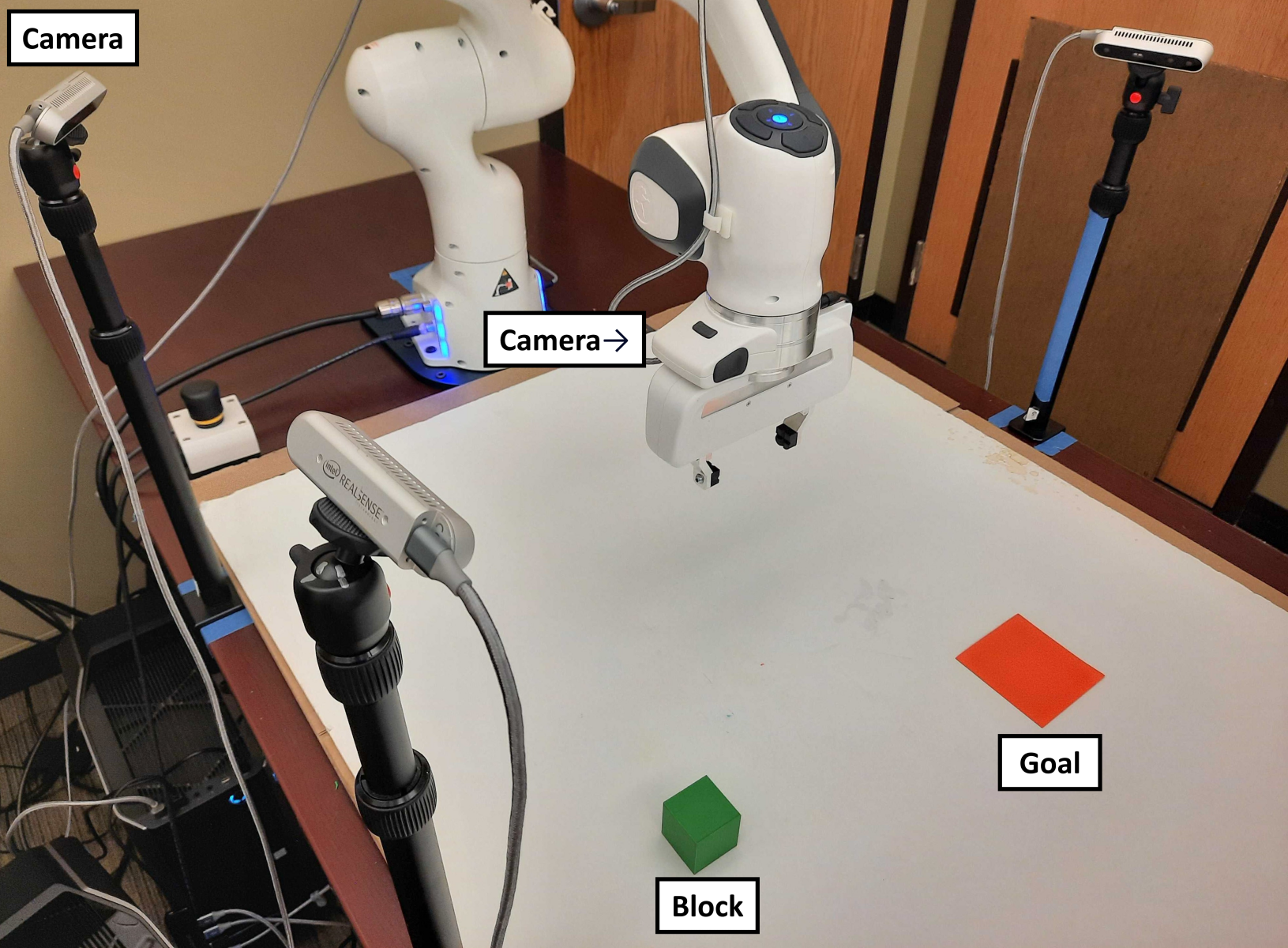}
  \caption{\label{fig:hardware}Hardware validation setup. A Franka Emika Panda arm was used to train a behavior cloning agent to move a 4 cm block to a goal location, shown with an 8 cm square orange piece of paper. A wrist-mounted RealSense camera, along with two stationary cameras mounted around the arm (one is labeled, one is off-screen), was used to collect visual observations for the agent.}
  \label{figurelabel}
\end{figure}

In order to test the viability of our single demonstration behavior cloning methodology for use on hardware, we implemented the push task (planar pick-and-place) using a Franka-Emika Panda robot as our manipulator \cite{franka2021}, and Intel RealSense D415 RGB-D cameras (we only used the RGB data) to record observations. The goal of this task was to pick up a block placed on the table and move it to a goal location. To indicate the goal location, we used an 8 cm orange square (in simulation, this had been done with a transparent cube). Additionally, due to the size of our table, we used a slightly smaller action space (60 cm square instead of 70 cm square). An image of our hardware setup can be seen in Figure \ref{fig:hardware}. To run the experiment, we first collected a series of augmented demos using the Franka robot. Given a single demonstration trajectory (the same demonstration trajectory, collected in VR, that we used for the simulation experiments) and a randomized goal and start location, we used our augmentation method to generate new trajectories. The generated trajectories were then replayed by the Franka Robot using a proportional controller, and the states (images from the RealSense cameras, the current end-effector position, and the current gripper width) and actions were recorded. At the end of the demonstration replay, an operator indicated whether the task was successful, and unsuccessful demonstrations were discarded.

Once the augmented demos were collected, we trained a behavior cloning agent using the same method we used for the simulation experiments, except we used three cameras (an end-effector camera, a side camera, and a front camera), rather than two, to compensate for our cameras' more limited view due to mounting constraints. We trained with 50, 100, and 200 augmented demonstrations. The trained policies were evaluated on the same hardware, using our temporal ensembling strategy. The agent had an accuracy of 35\%, 70\%, and 90\% for 50, 100, and 200 augmented demos, respectively. These values are quite similar to the results we got for the same conditions in simulation, following the trend of increasing accuracy with an increased number of demonstrations, but with slightly worse performance (see Table 1). Although not conclusive, the similarity between the hardware and simulation results suggests that the conclusions we draw from experiments in simulation can be extended to hardware applications. 
 
\section{CONCLUSIONS}
Our results show that a single human demonstration can be used to learn a successful policy through behavior cloning, as long as a sufficient augmentation method is applied. We showed that even a naive augmentation method - applying a piece-wise linear transform to the recorded trajectory - can allow behavior cloning to succeed with only a single human demonstration. By collecting many generated demonstrations, disposing of the failures, and training on the successes, even a brittle augmentation method, such as the one we use, can be used to train a robust policy. Although the diversity of the demonstrations the agent is training on may be limited, a combination of a CVAE to improve generalization and action-chunking to mitigate the negative effects of out-of-distribution states can overcome this limitation, enabling the agent to train a successful policy in multiple tasks.

Additionally, our work introduced a novel temporal ensembling method for combining action chunks at inference time. This method, which uses the standard deviation of the predicted actions as a proxy for changes in dynamics that may render prior action choices incorrect, mitigates the issues encountered by weighted average ensembling when the action distribution is multi-modal. By incorporating this simple statistical heuristic into the ensembling method's weighted average, we were able to improve our accuracy on the stack task from 60.8\% to 78.4\%, almost halving our error rate. Although this method is vital to the performance of our single demonstration behavior cloning algorithm, it can be applied to any temporal aggregation agent, making it a valuable tool for the behavior cloning community.

The main limitation of our work is that the benefits of using a single human demonstration with augmentation, as opposed to simply collecting more human demonstrations, are directly tied to the relative cost of collecting demonstrations vs. executing a demonstration. If the cost of the human demonstration dominates (such as when training in simulation), then this method essentially allows a user to collect a nearly unlimited number of demonstrations for the cost of only one. However, if the cost of an agent autonomously completing a task is similar to that of a human demonstrating the task (such as if a human has to play an active role in robot operation during autonomous demonstration collection), then our method is less impactful. Moving forward, we hope to address this discrepancy by combining hardware and simulation training, using our single demonstration method to train a policy in simulation, and then (using the same single demonstration) fine-tuning the policy on hardware demonstrations.

\addtolength{\textheight}{-2cm}   


\bibliographystyle{IEEEtran}
\bibliography{ref}

\begin{thebibliography}{10}
\providecommand{\url}[1]{#1}
\csname url@samestyle\endcsname
\providecommand{\newblock}{\relax}
\providecommand{\bibinfo}[2]{#2}
\providecommand{\BIBentrySTDinterwordspacing}{\spaceskip=0pt\relax}
\providecommand{\BIBentryALTinterwordstretchfactor}{4}
\providecommand{\BIBentryALTinterwordspacing}{\spaceskip=\fontdimen2\font plus
\BIBentryALTinterwordstretchfactor\fontdimen3\font minus
  \fontdimen4\font\relax}
\providecommand{\BIBforeignlanguage}[2]{{%
\expandafter\ifx\csname l@#1\endcsname\relax
\typeout{** WARNING: IEEEtran.bst: No hyphenation pattern has been}%
\typeout{** loaded for the language `#1'. Using the pattern for}%
\typeout{** the default language instead.}%
\else
\language=\csname l@#1\endcsname
\fi
#2}}
\providecommand{\BIBdecl}{\relax}
\BIBdecl

\bibitem{Pomerleau-1989-15721}
D.~Pomerleau, ``Alvinn: An autonomous land vehicle in a neural network,'' in
  \emph{Proceedings of (NeurIPS) Neural Information Processing Systems},
  D.~Touretzky, Ed.\hskip 1em plus 0.5em minus 0.4em\relax Morgan Kaufmann,
  December 1989, pp. 305 -- 313.

\bibitem{bojarski2016end}
M.~Bojarski, D.~D. Testa, D.~Dworakowski, B.~Firner, B.~Flepp, P.~Goyal, L.~D.
  Jackel, M.~Monfort, U.~Muller, J.~Zhang, X.~Zhang, J.~Zhao, and K.~Zieba,
  ``End to end learning for self-driving cars,'' 2016.

\bibitem{8855753}
W.~Farag and Z.~Saleh, ``Behavior cloning for autonomous driving using
  convolutional neural networks,'' in \emph{2018 International Conference on
  Innovation and Intelligence for Informatics, Computing, and Technologies
  (3ICT)}, 2018, pp. 1--7.

\bibitem{vinyals2019grandmaster}
O.~Vinyals, I.~Babuschkin, W.~M. Czarnecki, M.~Mathieu, A.~Dudzik, J.~Chung,
  D.~H. Choi, R.~Powell, T.~Ewalds, P.~Georgiev \emph{et~al.}, ``Grandmaster
  level in starcraft ii using multi-agent reinforcement learning,''
  \emph{Nature}, vol. 575, no. 7782, pp. 350--354, 2019.

\bibitem{torabi2018behavioral}
F.~Torabi, G.~Warnell, and P.~Stone, ``Behavioral cloning from observation,''
  \emph{arXiv preprint arXiv:1805.01954}, 2018.

\bibitem{Mandlekar2021WhatMI}
\BIBentryALTinterwordspacing
A.~Mandlekar, D.~Xu, J.~Wong, S.~Nasiriany, C.~Wang, R.~Kulkarni, L.~Fei-Fei,
  S.~Savarese, Y.~Zhu, and R.~Mart'in-Mart'in, ``What matters in learning from
  offline human demonstrations for robot manipulation,'' in \emph{Conference on
  Robot Learning}, 2021. [Online]. Available:
  \url{https://api.semanticscholar.org/CorpusID:236956615}
\BIBentrySTDinterwordspacing

\bibitem{shafiullah2022behavior}
N.~M. Shafiullah, Z.~Cui, A.~A. Altanzaya, and L.~Pinto, ``Behavior
  transformers: Cloning $k$ modes with one stone,'' \emph{Advances in neural
  information processing systems}, vol.~35, pp. 22\,955--22\,968, 2022.

\bibitem{9289148}
J.~Choi, H.~Kim, Y.~Son, C.-W. Park, and J.~H. Park, ``Robotic behavioral
  cloning through task building,'' in \emph{2020 International Conference on
  Information and Communication Technology Convergence (ICTC)}, 2020, pp.
  1279--1281.

\bibitem{ross2011}
S.~Ross, G.~Gordon, and D.~Bagnell, ``A reduction of imitation learning and
  structured prediction to no-regret online learning,'' in \emph{Proceedings of
  the fourteenth international conference on artificial intelligence and
  statistics}.\hskip 1em plus 0.5em minus 0.4em\relax JMLR Workshop and
  Conference Proceedings, 2011, pp. 627--635.

\bibitem{Tu2021OnTS}
\BIBentryALTinterwordspacing
S.~Tu, A.~Robey, T.~Zhang, and N.~Matni, ``On the sample complexity of
  stability constrained imitation learning,'' in \emph{Conference on Learning
  for Dynamics \& Control}, 2021. [Online]. Available:
  \url{https://api.semanticscholar.org/CorpusID:235358617}
\BIBentrySTDinterwordspacing

\bibitem{10.1145/3054912}
\BIBentryALTinterwordspacing
A.~Hussein, M.~M. Gaber, E.~Elyan, and C.~Jayne, ``Imitation learning: A survey
  of learning methods,'' \emph{ACM Comput. Surv.}, vol.~50, no.~2, apr 2017.
  [Online]. Available: \url{https://doi.org/10.1145/3054912}
\BIBentrySTDinterwordspacing

\bibitem{lai2022action}
L.~Lai, A.~Z. Huang, and S.~J. Gershman, ``Action chunking as policy
  compression,'' 2022.

\bibitem{8971711}
S.-J. Lee, T.~Y. Chun, H.~W. Lim, and S.-H. Lee, ``Path tracking control using
  imitation learning with variational auto-encoder,'' in \emph{2019 19th
  International Conference on Control, Automation and Systems (ICCAS)}, 2019,
  pp. 501--505.

\bibitem{wang2023diffusion}
H.-C. Wang, S.-F. Chen, M.-H. Hsu, C.-M. Lai, and S.-H. Sun, ``Diffusion
  model-augmented behavioral cloning,'' 2023.

\bibitem{8461249}
T.~Zhang, Z.~McCarthy, O.~Jow, D.~Lee, X.~Chen, K.~Goldberg, and P.~Abbeel,
  ``Deep imitation learning for complex manipulation tasks from virtual reality
  teleoperation,'' in \emph{2018 IEEE International Conference on Robotics and
  Automation (ICRA)}, 2018, pp. 5628--5635.

\bibitem{10161119}
A.~George, A.~Bartsch, and A.~B. Farimani, ``Minimizing human assistance:
  Augmenting a single demonstration for deep reinforcement learning,'' in
  \emph{2023 IEEE International Conference on Robotics and Automation (ICRA)},
  2023, pp. 5027--5033.

\bibitem{zhao2023learning}
T.~Z. Zhao, V.~Kumar, S.~Levine, and C.~Finn, ``Learning fine-grained bimanual
  manipulation with low-cost hardware,'' 2023.

\bibitem{ratliff2007}
N.~Ratliff, J.~A. Bagnell, and S.~S. Srinivasa, ``Imitation learning for
  locomotion and manipulation,'' in \emph{2007 7th IEEE-RAS International
  Conference on Humanoid Robots}.\hskip 1em plus 0.5em minus 0.4em\relax IEEE,
  2007, pp. 392--397.

\bibitem{nakanishi2004}
J.~Nakanishi, J.~Morimoto, G.~Endo, G.~Cheng, S.~Schaal, and M.~Kawato,
  ``Learning from demonstration and adaptation of biped locomotion,''
  \emph{Robotics and autonomous systems}, vol.~47, no. 2-3, pp. 79--91, 2004.

\bibitem{reichlin2022back}
A.~Reichlin, G.~L. Marchetti, H.~Yin, A.~Ghadirzadeh, and D.~Kragic, ``Back to
  the manifold: Recovering from out-of-distribution states,'' in \emph{2022
  IEEE/RSJ International Conference on Intelligent Robots and Systems
  (IROS)}.\hskip 1em plus 0.5em minus 0.4em\relax IEEE, 2022, pp. 8660--8666.

\bibitem{pmlr-v80-kang18a}
\BIBentryALTinterwordspacing
B.~Kang, Z.~Jie, and J.~Feng, ``Policy optimization with demonstrations,'' in
  \emph{Proceedings of the 35th International Conference on Machine Learning},
  ser. Proceedings of Machine Learning Research, J.~Dy and A.~Krause, Eds.,
  vol.~80.\hskip 1em plus 0.5em minus 0.4em\relax PMLR, 10--15 Jul 2018, pp.
  2469--2478. [Online]. Available:
  \url{https://proceedings.mlr.press/v80/kang18a.html}
\BIBentrySTDinterwordspacing

\bibitem{subramanian2016exploration}
K.~Subramanian, C.~L. Isbell~Jr, and A.~L. Thomaz, ``Exploration from
  demonstration for interactive reinforcement learning,'' in \emph{Proceedings
  of the 2016 international conference on autonomous agents \& multiagent
  systems}, 2016, pp. 447--456.

\bibitem{brys2015reinforcement}
T.~Brys, A.~Harutyunyan, H.~B. Suay, S.~Chernova, M.~E. Taylor, and
  A.~Now{\'e}, ``Reinforcement learning from demonstration through shaping,''
  in \emph{Twenty-fourth international joint conference on artificial
  intelligence}, 2015.

\bibitem{10.1145/1553374.1553380}
\BIBentryALTinterwordspacing
Y.~Bengio, J.~Louradour, R.~Collobert, and J.~Weston, ``Curriculum learning,''
  in \emph{Proceedings of the 26th Annual International Conference on Machine
  Learning}, ser. ICML '09.\hskip 1em plus 0.5em minus 0.4em\relax New York,
  NY, USA: Association for Computing Machinery, 2009, p. 41–48. [Online].
  Available: \url{https://doi.org/10.1145/1553374.1553380}
\BIBentrySTDinterwordspacing

\bibitem{salimans2018learning}
T.~Salimans and R.~Chen, ``Learning montezuma's revenge from a single
  demonstration,'' 2018.

\bibitem{schulman2017proximal}
J.~Schulman, F.~Wolski, P.~Dhariwal, A.~Radford, and O.~Klimov, ``Proximal
  policy optimization algorithms,'' \emph{arXiv preprint arXiv:1707.06347},
  2017.

\bibitem{florensa2018reverse}
C.~Florensa, D.~Held, M.~Wulfmeier, M.~Zhang, and P.~Abbeel, ``Reverse
  curriculum generation for reinforcement learning,'' 2018.

\bibitem{lipton2016}
Z.~C. Lipton, J.~Gao, L.~Li, X.~Li, F.~Ahmed, and L.~Deng, ``Efficient
  exploration for dialog policy learning with deep bbq networks \& replay
  buffer spiking,'' \emph{CoRR abs/1608.05081}, 2016.

\bibitem{nair2018}
A.~Nair, B.~McGrew, M.~Andrychowicz, W.~Zaremba, and P.~Abbeel, ``Overcoming
  exploration in reinforcement learning with demonstrations,'' in \emph{2018
  IEEE international conference on robotics and automation (ICRA)}.\hskip 1em
  plus 0.5em minus 0.4em\relax IEEE, 2018, pp. 6292--6299.

\bibitem{vargas2019creativity}
J.~C. Vargas, M.~Bhoite, and A.~B. Farimani, ``Creativity in robot manipulation
  with deep reinforcement learning,'' 2019.

\bibitem{46201}
\BIBentryALTinterwordspacing
A.~Vaswani, N.~Shazeer, N.~Parmar, J.~Uszkoreit, L.~Jones, A.~N. Gomez,
  L.~Kaiser, and I.~Polosukhin, ``Attention is all you need,'' 2017. [Online].
  Available: \url{https://arxiv.org/pdf/1706.03762.pdf}
\BIBentrySTDinterwordspacing

\bibitem{bharadhwaj2023roboagent}
H.~Bharadhwaj, J.~Vakil, M.~Sharma, A.~Gupta, S.~Tulsiani, and V.~Kumar,
  ``Roboagent: Generalization and efficiency in robot manipulation via semantic
  augmentations and action chunking,'' 2023.

\bibitem{george2023openvr}
A.~George, A.~Bartsch, and A.~B. Farimani, ``Openvr: Teleoperation for
  manipulation,'' 2023.

\bibitem{coumans2019}
E.~Coumans and Y.~Bai, ``Pybullet, a python module for physics simulation for
  games, robotics and machine learning,'' \url{http://pybullet.org},
  2016--2019.

\bibitem{dikshit2023robochop}
A.~Dikshit, A.~Bartsch, A.~George, and A.~B. Farimani, ``Robochop: Autonomous
  framework for fruit and vegetable chopping leveraging foundational models,''
  2023.

\bibitem{gallouedec2021}
Q.~Gallou{\'e}dec, N.~Cazin, E.~Dellandr{\'e}a, and L.~Chen, ``Multi-goal
  reinforcement learning environments for simulated franka emika panda robot,''
  \emph{arXiv preprint arXiv:2106.13687}, 2021.

\bibitem{franka2021}
\emph{Franka Emika Robot's Instruction Handbook}.\hskip 1em plus 0.5em minus
  0.4em\relax Franka Emika GmbH, 2021.

\end{thebibliography}

\end{document}